\pdfoutput=1

\documentclass[11pt]{article}

\usepackage[final]{acl}

\usepackage{times}
\usepackage{latexsym}
\usepackage{multirow}
\usepackage{graphicx}
\usepackage{booktabs}
\usepackage{enumitem}
\usepackage{algorithm}
\usepackage{algorithmic}
\usepackage{amsmath}
\usepackage{makecell}
\usepackage{bm}
\usepackage{colortbl}
\usepackage{subcaption}
\usepackage{tabularx}
\usepackage{pifont}
\newcommand{\cmark}{\ding{51}} 
\newcommand{\xmark}{\ding{55}} 
\usepackage{siunitx}     
\usepackage[T1]{fontenc}

\usepackage[utf8]{inputenc}

\usepackage{microtype}

\usepackage{inconsolata}
\usepackage{pifont}

\def\method{\textsc{FinMME}}
\title{
\method{}: Benchmark Dataset for Financial \\ Multi-Modal Reasoning Evaluation
}

\author{
Junyu Luo\textsuperscript{1,2}\thanks{Equal contribution.}, 
Zhizhuo Kou\textsuperscript{3}\footnotemark[1], 
Liming Yang\textsuperscript{2}\footnotemark[1], 
Xiao Luo\textsuperscript{4},
Jinsheng Huang\textsuperscript{1,2},\\
\textbf{Zhiping Xiao}\textsuperscript{5}\footnotemark[2],
\textbf{Jingshu Peng}\textsuperscript{3},
\textbf{Chengzhong Liu}\textsuperscript{3},
\textbf{Jiaming Ji}\textsuperscript{3},\\
\textbf{Xuanzhe Liu}\textsuperscript{2},
\textbf{Sirui Han}\textsuperscript{3}\thanks{Corresponding author.},
\textbf{Ming Zhang}\textsuperscript{1,2}\footnotemark[2],
\textbf{Yike Guo}\textsuperscript{3},\\
{\textsuperscript{1} State Key Laboratory for Multimedia Information Processing, PKU-Anker LLM Lab} \\
{\textsuperscript{2} School of Computer Science, Peking University} \quad
{\textsuperscript{3} HKUST} \\
{\textsuperscript{4} University of California, Los Angeles} \quad
{\textsuperscript{5} University of Washington}\\
{\tt Dataset: \url{https://huggingface.co/datasets/luojunyu/FinMME}}\\
}

\def\ie{\emph{i.e}.}

\def\etc{\emph{etc}.}

\definecolor{LightCyan}{rgb}{0.88,1,1}

\newcommand{\paratitle}[1]{\noindent\emph{\textbf{#1}}}

\usepackage{amsmath,amsfonts,bm}
\usepackage{algorithm,amsmath,amssymb,bm,amsthm}
\usepackage{algorithmic}
\usepackage{enumitem}

\newtheorem*{lemma*}{Lemma}

\def\eqref#1{equation~\ref{#1}}

\def\1{\bm{1}}

\DeclareMathAlphabet{\mathsfit}{\encodingdefault}{\sfdefault}{m}{sl}
\SetMathAlphabet{\mathsfit}{bold}{\encodingdefault}{\sfdefault}{bx}{n}

\begin{document}
\maketitle
\begin{abstract}

Multimodal Large Language Models (MLLMs) have experienced rapid development in recent years. However, in the financial domain, there is a notable lack of effective and specialized multimodal evaluation datasets. To advance the development of MLLMs in the finance domain, we introduce \method{}, encompassing more than 11,000 high-quality financial research samples across 18 financial domains and 6 asset classes, featuring 10 major chart types and 21 subtypes. We ensure data quality through 20 annotators and carefully designed validation mechanisms. Additionally, we develop FinScore, an evaluation system incorporating hallucination penalties and multi-dimensional capability assessment to provide an unbiased evaluation. Extensive experimental results demonstrate that even state-of-the-art models like GPT-4o exhibit unsatisfactory performance on \method{}, highlighting its challenging nature. The benchmark exhibits high robustness with prediction variations under different prompts remaining below 1\%, demonstrating superior reliability compared to existing datasets. Our dataset and evaluation protocol are available at  \url{https://github.com/luo-junyu/FinMME}.

\end{abstract}

\section{Introduction}

Multimodal Large Language Models (MLLMs) have demonstrated remarkable progress in data comprehension and understanding~\cite{fu2024mmesurvey}, with their capabilities being evaluated through various benchmarks such as MME~\cite{fu2024mme}, SEED~\cite{li2024seed}, MMC~\cite{liu2023mmc}, MMMU~\cite{yue2024mmmu}. The establishment of effective datasets and benchmarks has been instrumental in guiding model optimization and comparative analysis, significantly accelerating the development of multimodal large models.

The financial domain\cite{chen2022convfinqa,li2023large}, characterized by its knowledge-intensive nature and rich multimodal data, presents an ideal application space for MLLMs, particularly in areas such as research report analysis~\cite{zhao2024revolutionizing}, risk forecasting~\cite{sawhney2020multimodal}, and market analysis~\cite{liu2024large}. However, the financial sector poses unique challenges due to its inherent complexity, higher data and knowledge density, and extensive domain expertise requirements, necessitating specialized domain-specific evaluation frameworks. Despite this need, there is currently a notable absence of comprehensive, high-quality multimodal datasets specifically designed for evaluating and optimizing MLLMs in the financial domain.

It is non-trivial to design a comprehensive, high-quality financial multimodal dataset, which presents several fundamental challenges:
\vspace{-2mm}
\begin{itemize}[leftmargin=*]
    \item \textbf{Data Volume:} Limited volume could lead to high variance in results and limited stability.
    \vspace{-2mm}
    \item \textbf{Data Quality:} MLLM annotated datasets may introduce hallucination-based errors. Moreover, high-knowledge-density financial multimodal datasets remain notably underexplored.
    \vspace{-2mm}
    \item \textbf{Domain-specificity and Difficulty:} While MLLMs achieve 80-90\% accuracy on general benchmarks~\cite{masry2022chartqa,li2023seed,liu2024mmbench}, financial tasks require both higher accuracy and domain expertise, demanding more rigorous evaluation scenarios.
\end{itemize}

To address these challenges, we introduce \method{}, a comprehensive and high-quality financial multimodal dataset with the following key features: \ding{182} \textbf{Comprehensive Financial Knowledge Coverage}: \method{} incorporates more than 11,000 rigorously selected financial samples spanning 18 core domains and 6 asset classes. Each sample contains financial charts (10 major types with 21 subtypes), professional research descriptions, hierarchical metadata, and QA annotations, reflecting real-world financial analysis workflows. \ding{183} \textbf{High Data Quality}: We employed 20 annotators and implemented carefully designed validation mechanisms, maintaining annotation error rates below 1\% for critical questions. \ding{184} \textbf{Innovative Quality Control}: We leverage MLLMs' external consistency to enhance annotation quality and efficiency, with expert review for cases where multiple models and human annotators disagree. \ding{185} \textbf{Novel Evaluation Metrics}: We introduce a hierarchical evaluation framework encompassing comprehensive perception, fine-grained analysis, and cognitive reasoning. Additionally, we designed FinScore, which provides unbiased evaluation across multiple financial domains while incorporating hallucination penalties to address the financial sector's low tolerance for inaccuracies. \ding{186} \textbf{Challenge and Effectiveness}: Extensive experiments on \method{} demonstrate that even leading MLLMs (GPT-4o, Germini Flash and Claude 3.5 Sonnet) achieve just over 50\% performance, highlighting the significant challenges and necessity for multimodal research in the financial domain. We tested 6 proprietary models and 11 open-source models, the prediction standard deviation across different prompts remains below 1\%, confirming \method{}'s robustness.

In summary, \method{} establishes a new benchmark for financial MLLMs through its comprehensive data coverage, rigorous annotation process, and hierarchical evaluation framework, advancing multimodal capabilities in specialized financial applications.

\section{Related Work}

\subsection{Multi-modal Large Language Models}

Recent advances in Multi-modal Large Language Models~(MLLMs) have demonstrated remarkable capabilities in unified visual-linguistic understanding as agentic AI~\cite{agentsurvey2025}, with open-source models like QwenVL~\cite{bai2023qwen}, Vita~\cite{fu2024vita}, VILA~\cite{lin2024vila}, CogVLM~\cite{wang2023cogvlm}, and LLaVA~\cite{li2024llava}, alongside proprietary models including GPT-4o\footnote{https://openai.com/index/hello-gpt-4o/}, Claude 3.5 Sonnet\footnote{https://www.anthropic.com/news/claude-3-5-sonnet} and Gemini~\cite{team2023gemini} showing strong performance in general domain tasks. However, despite their sophisticated encoder-decoder architectures for cross-modal understanding, our evaluation reveals that these MLLMs significantly underperform in knowledge-intensive financial tasks, highlighting the need for specialized datasets such as \method{} to advance financial MLLMs.

\begin{table}[t]
\centering
\setlength{\tabcolsep}{5pt}
\resizebox{\linewidth}{!}{%
\begin{tabular}{lcccc}
\toprule
\textbf{Dataset} &  \makecell{\textbf{Dataset}\\\textbf{Volume}} & \makecell{\textbf{Human}\\\textbf{Anno.}} & \makecell{\textbf{Specific}\\\textbf{Domain}} & \makecell{\textbf{GPT-4o}\\\textbf{Performance}} \\

\midrule
MMStar       & 1500   & \xmark    & \xmark   & 62 \\
MM-Vet       & 218    & \xmark    & \xmark   & 72 \\
MME          & 2374   & \cmark    & \xmark   & -- \\
MMBench      & 3217   & \cmark    & \xmark   & 83 \\
MMC          & 2126   & \cmark    & \xmark   & 76 \\
MMMU~(Full) & 11550   & \cmark    & \xmark   & 63 \\
\midrule
MMMU~(Finance) & 390   & \cmark    & Finance   & - \\
MME-Finance  & 1171   & \xmark    & Finance  & 63 \\
\textbf{\method{}}~(Ours) & 11099 & \cmark & Finance  & 47 \\
\bottomrule
\end{tabular}
}
\caption{
Comparison with existing benchmarks. \method{} provides a comprehensive and high-quality dataset for the financial multimodal domain.
}

\vspace{-2mm}
\label{tab:multimodal-datasets}
\end{table}

\subsection{Multi-Modal Evaluation Datasets}

Recent advancements in MLLMs have demonstrated exceptional capabilities across a wide array of complex tasks, including MMStar~\cite{chen2024we}, MM-Vet~\cite{yu2023mm}, MME~\cite{fu2024mme}, MMBench~\cite{liu2024mmbench}, MMC~\cite{liu2023mmc}, MMMU~\cite{yue2024mmmu}, and others~\cite{li2023seed,li2024seed,huang2024mmevalpro}. Comprehensive benchmarks are essential not only to gauge progress in general multimodal reasoning but also to pinpoint areas that require further refinement. However, domain-specific evaluation remains limited, particularly in the finance domain, where the high knowledge density and inherent complexity of financial data demand specialized evaluation frameworks. More related background can be found in Appendix~\ref{appendix:related_work}.

\paratitle{Differences from Existing Datasets}
As shown in Table~\ref{tab:multimodal-datasets}, existing multimodal benchmarks are constrained by data scale, annotation quality, domain coverage and task complexity\footnote{The performance is from official reports or quoted~\cite{fu2025vita}. MMMU~(Finance) is the domain-specific subset.}. While the concurrent work MME-Finance~\cite{gan2024mme} also targets financial multimodal evaluation, it faces limitations in data volume and annotation quality. In contrast, \method{} offers a comprehensive, high-quality large-scale dataset specifically designed for financial multimodal tasks. We provide a detailed comparison with existing financial domain datasets in Appendix~\ref{appendix:dataset_comparison} to highlight the advantages.

\begin{figure*}[t]
    \centering
    \includegraphics[width=\textwidth]{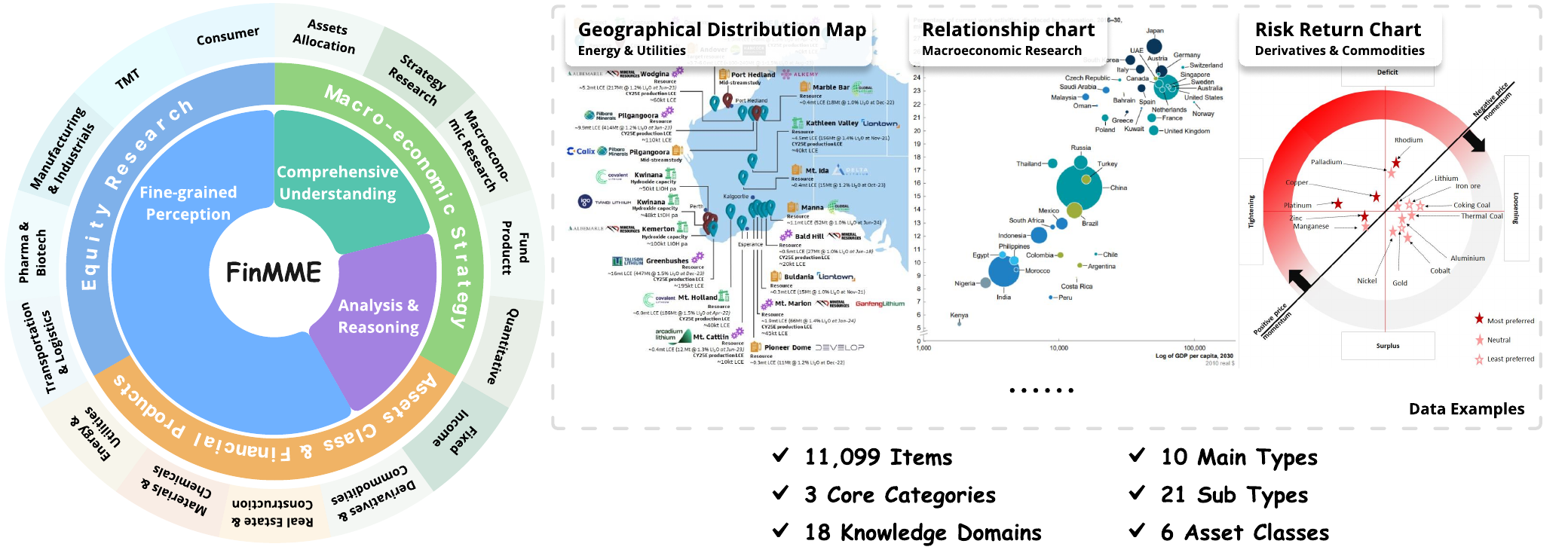}
    \caption{The Comprehensive Taxonomy, Data Examples and Statistical Characteristics of \method{}. The circular taxonomy diagram shows three core cognitive levels, knowledge categories and domains.}
    \label{fig:finmme-taxo}
    \vspace{-4mm}
\end{figure*}

\section{\method{} Dataset: High-Quality Financial Multi-Modal Dataset}


\method{} comprises more than $11,000$ high-quality financial multi-modal samples, with each sample consisting of multi-modal metadata and question information. 
All data undergoes a rigorous quality control process to ensure reliability.
Our dataset design was informed by discussions with six financial domain experts (detailed consultation records in Appendix~\ref{appendix:expert_consultation}).
This section provides a comprehensive introduction to the FinMME dataset, including detailed data classification and statistics~(Section~\ref{sec:data_classification}), question-answer design~(Section~\ref{sec:question_answer_design}), data sources~(Section~\ref{sec:data_sources}), annotation process~(Section~\ref{sec:annotation_process}), and quality control protocols~(Section~\ref{sec:quality_control}).

\begin{figure*}[t]
    \centering
    \includegraphics[width=\textwidth]{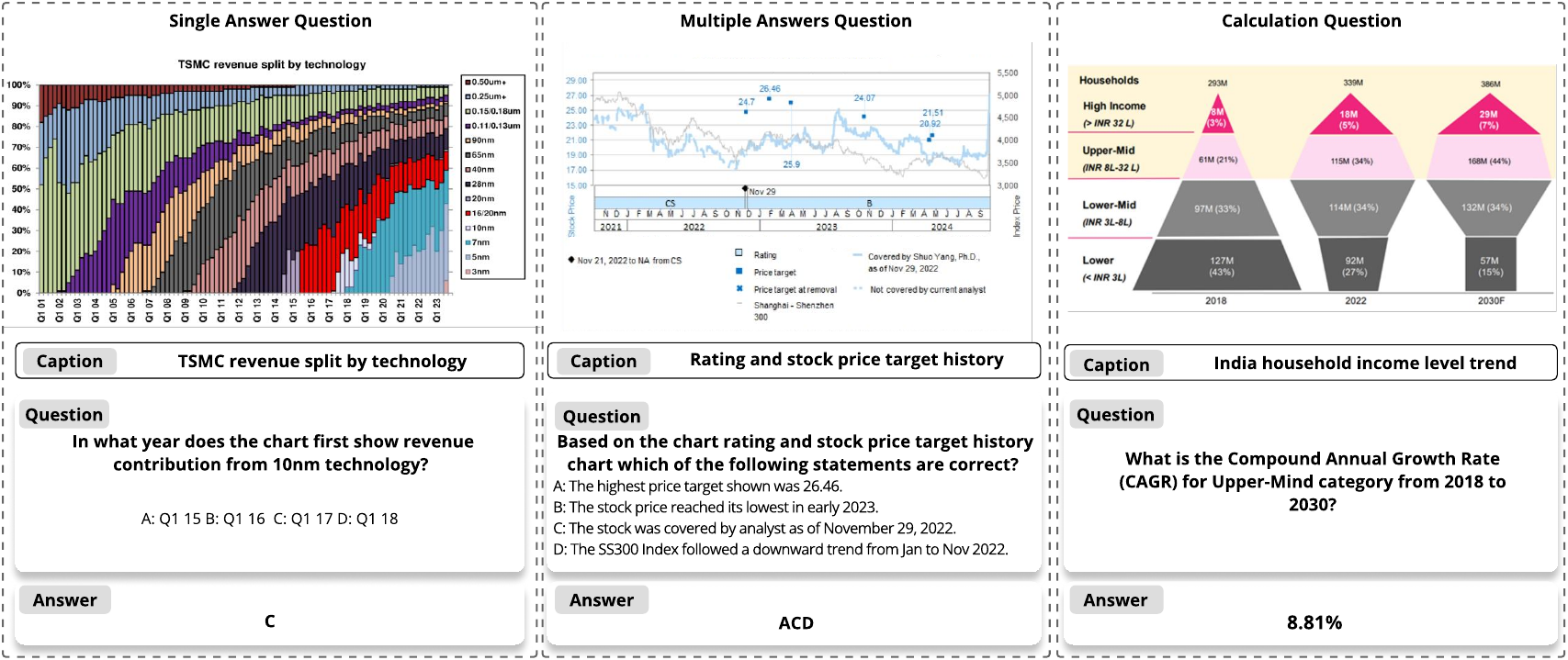}
    \vspace{-2mm}
    \caption{Representative examples of different question types in \method{} dataset.}
    \vspace{-4mm}
    \label{fig:finmme-example-qa}
\end{figure*}

%

\subsection{Statistical Characteristics}\label{sec:data_classification}

The multi-modal metadata encompasses financial images, image captions, professional research report descriptions, and fine-grained data labels~(\ie, target markets, asset classes, and detailed data class labels). 
The question information includes problem statements, multiple choice options, standard answers (with unit and error tolerance ranges for calculation questions), and question type labels.
Dataset statistics are shown in Table~\ref{tab:dataset-statistics}.

\subsection{Fine-grained Data Labels}

\paratitle{Knowledge Domain.} \method{} aims to provide comprehensive coverage of financial knowledge domains, encompassing 18 core financial domains: TMT~(Technology, Media \& Telecom), Consumer, Pharmaceuticals \& Biotechnology, Financials, Real Estate \& Construction, Industrials \& Manufacturing, Energy \& Utilities, Materials \& Chemicals, Military \& Defense, Transportation \& Logistics, Macroeconomic Research, Strategy Research, Broad Asset Allocation, Equity Research, Fixed Income, Fixed Income Quantitative, Derivatives \& Commodities, and Fund Products. The taxonomy is in Figure~\ref{fig:finmme-taxo}. This extensive coverage effectively reflects the modern financial knowledge system.

\begin{table}[t]
\centering
\renewcommand{\arraystretch}{1}
\begin{tabular}{@{}lr@{}}
    \toprule
    \textbf{Statistic} & \textbf{Number} \\
    \midrule
    \multicolumn{2}{@{}l}{\textit{Dataset Overview}} \\
    \cmidrule(l){1-2}
    \textbf{Total Samples} & 11,099 \\
    \midrule
    \multicolumn{2}{@{}l}{\textit{Cognitive Level Distribution}} \\
    \cmidrule(l){1-2}
    Comprehensive Understanding & 2,333 \\
    Fine-grained Perception & 6,466 \\
    Analysis and Reasoning & 2,300 \\
    \midrule
    \multicolumn{2}{@{}l}{\textit{Core Knowledge Domain}} \\
    \cmidrule(l){1-2}
    Equity Research & 7,601 \\
    Macroeconomic Research & 1,485 \\
    Assets Class and Financial Products & 2,013 \\
    \midrule
    Unique Images & 4,458 \\
    Average Question Length & 24.1 \\
    Average Caption Length & 10.8 \\
    \bottomrule
\end{tabular}
\vspace{-2mm}
\caption{Statistical characteristics of the \method{} dataset, including question types, cognitive levels, and knowledge domains.}
\vspace{-4mm}
\label{tab:dataset-statistics}
\end{table}

\paratitle{Data Class.} FinMME incorporates diverse data classes, categorized into 10 main classes and 21 subclasses. The main classes comprise Time Series, Distribution Charts, Proportional Charts, Relationship Charts, Financial Reports, Risk Analysis, Market Structure, Geographical Charts, Process Flow, \etc. To facilitate future research, we have meticulously annotated each image with both main class and subclass categories, with details provided in  Appendix~\ref{appendix:dataset}.

\paratitle{Asset Class.} We effectively differentiate the multi-modal data according to $6$ asset classes to support cross-asset analysis. The dataset covers Equity, Foreign Exchange, Rates, Commodity, Credits, and Cross-Asset. These asset class labels enable targeted model evaluation across different market segments and facilitate the assessment of specialized knowledge in distinct financial instruments.

\begin{figure*}[t]
    \centering
    \includegraphics[width=\textwidth]{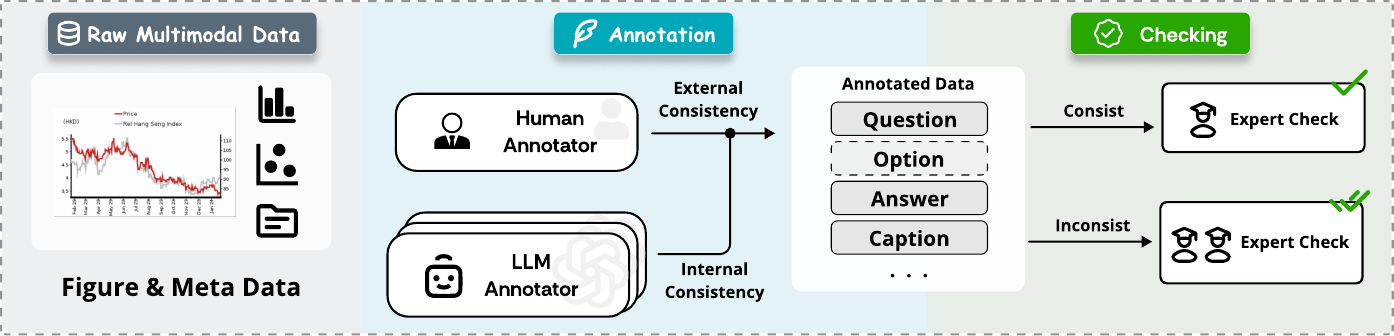}
    \caption{The annotation pipeline of \method{}. The process consists of three main stages: (1) Raw Multimodal Data collection, (2) Annotation through parallel human and LLM annotators to ensure external and internal consistency, and (3) Quality Control checking where expert reviewers validate consistent annotations and resolve inconsistencies.}
    \label{fig:finmme-anno}
\end{figure*}

\subsection{Question-Answer Design}\label{sec:question_answer_design}

We establish a hierarchical evaluation framework to comprehensively assess MLLMs' capabilities in the financial domain. This framework encompasses three fundamental dimensions:

\paratitle{Comprehensive Perception.} This dimension evaluates models' ability to perform temporal sequence recognition, horizontal comparisons, holistic discrimination, and multi-chart analysis. The assessment is primarily conducted through multiple-choice questions~ (single answer and multiple answers), focusing on models' capacity to comprehend and interpret complex financial visualizations and their interrelationships.

\paratitle{Fine-grained Perception.} This aspect examines numerical extraction and local variation analysis capabilities. The evaluation utilizes multiple-choice questions~(single answer and multiple answers) to assess models' precision in identifying and analyzing specific data points and localized patterns within financial contexts.

\paratitle{Cognition and Reasoning.} This dimension encompasses data inference, cross-modal understanding, trend prediction, causal analysis, scenario-based decision support, and hypothesis analysis. The assessment combines computational problems and multiple-choice questions to evaluate models' advanced reasoning capabilities in financial scenarios, including their ability to synthesize information across modalities and make informed predictions.

\subsection{Data Sources}\label{sec:data_sources}

Adhering to compliance principles, we collected over $7,000$ professional research reports and web page screenshots through a hybrid approach combining manual curation and automated crawling, from which we extracted high-quality financial images and associated text. Throughout the collection process, we prioritized copyright compliance and selected materials authorized for public dissemination. All data underwent a rigorous three-stage cleaning process: automated deduplication, format standardization, and manual review, ensuring the authoritativeness and legality of data sources.

\subsection{Annotation Process}\label{sec:annotation_process}

\paratitle{Annotation Team.} 
We recruited a team of $20$ annotators, consisting of $12$ Junior annotators and $8$ Experts. 
Junior annotators with basic finance knowledge were responsible for question review, reformulation, and independent problem-solving. 
The Expert group included (i) $4$ people from academia specializing in STEM and finance, holding at least a master's degree, and (ii) $4$ finance industry professionals. These experts were tasked with dataset question selection, quality assessment, and answer verification. 

\paratitle{Time Investment.} The annotation and review process required approximately $800$ cumulative hours of work from the $20$-member team, with time estimates aggregated from individual contributions.

\subsection{Quality Control Protocol}\label{sec:quality_control}
We designed an innovative quality control methodology, as illustrated in Figure \ref{fig:finmme-anno}. While ensuring dataset quality, we use LLMs to achieve a more efficient dataset construction process through a three-stage pipeline. First, we collect and prepare the raw multimodal data. Second, in the annotation stage, we employ a parallel annotation strategy where both human annotators and multiple LLM annotators independently process the data. This dual-track approach helps establish both external consistency (through human annotations) and internal consistency (through multiple LLM predictions). The annotated data includes questions, options, answers, captions, and other relevant metadata. Finally, in the quality control stage, we implement a consistency-based review process: when human and LLM annotations align, a single expert performs a validation check; when discrepancies occur, multiple experts conduct a thorough review to determine the final ground truth. This systematic approach ensures high-quality annotations while optimizing the efficiency of expert involvement.

\subsection{Summary}\label{sec:dataset_summary}

\method{} distinguishes itself from existing datasets through three key characteristics: superior quality, comprehensive coverage, and fine-grained label annotations. The dataset features high-quality multi-modal data spanning diverse financial knowledge domains, accompanied by meticulously annotated classifications and question-answer pairs. These distinctive attributes enable effective evaluation of MLLMs' performance in complex financial scenarios. The combination of the above positions \method{} as a robust benchmark for assessing multi-modal language models' capabilities in professional financial applications.

\begin{figure}[t]
    \centering
    \includegraphics[width=\linewidth]{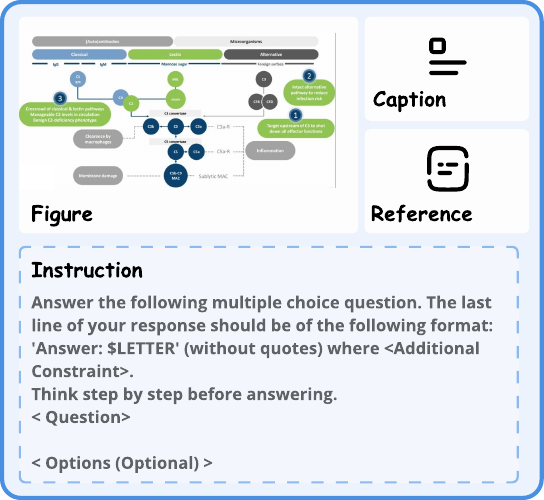}
    \caption{Illustration of the evaluation prompt template. }
    \label{fig:finmme-prompt}
\end{figure}

\section{\method{} Benchmark: Comprehensive Financial Multi-Modal Evaluation}\label{sec:benchmark}

To ensure comprehensive evaluation, we employ a combination of multiple-choice questions~(MCQs) and computational problems. The MCQs include both single-answer and multiple-answer formats, with an increased emphasis on multiple-answer questions compared to existing datasets. This design choice aims to better challenge models and reduce hallucination tendencies, as multiple-answer questions require more precise understanding and exhibit lower tolerance for incorrect selections.

\subsection{Hallucination Penalty}\label{sec:hallucination_penalty}

For multiple-answer questions, we introduce a scoring mechanism that effectively balances reward for correct answers with penalties for over-selection. The raw score for a single multiple-choice question is calculated as:
\begin{equation}\label{eq:penalty_score}
    S_q = \max(0, \frac{c}{n} - \frac{i}{s})\,,
\end{equation}
where $S_q$ represents the raw score for a single multiple-choice question, $c$ is the number of correct selections, $n$ is the total number of options, $i$ is the number of incorrect selections, and $s$ is the total selections made by the model. This formulation penalizes hallucination by reducing scores proportionally to incorrect selection ratios while normalizing based on the total options available.

\subsection{Knowledge-unbiased Evaluation}\label{sec:domain_unbiased_score}

Financial knowledge domains inherently vary in complexity and difficulty. For instance, quantitative analysis in derivatives typically presents greater challenges than basic equity research. To address these variations and ensure fair evaluation, we implement domain-normalized scoring:
\begin{equation}\label{eq:domain_normalized_score}
    F = \frac{1}{K}\sum_{k=1}^K \frac{1}{N_k}\sum_{i=1}^{N_k} S_{k,i}\,,
\end{equation}
where $S_{k,i}$ represents the score of the $i$-th question in domain $k$, $N_k$ is the total number of questions in domain $k$, and $K$ is the total number of domains. This formulation first calculates the average performance within each domain, then takes the mean across all domains, ensuring each knowledge domain contributes equally to the final score regardless of its number of questions.

\subsection{FinScore}\label{sec:finscore}

Financial applications demand both high accuracy and low hallucination due to the critical nature of investment decisions. To address this dual requirement, we introduce FinScore ($\mathcal{F}$) that combines domain-normalized performance with hallucination penalties, reflecting a model's practical value in financial contexts.

We first define the hallucination penalty rate $P_H$, which represents the average ratio of incorrect selections across the dataset:
\begin{equation}\label{eq:hallucination_penalty_rate}
P_H = \text{mean}\left(\frac{i}{s}\right)\,,
\end{equation}
where the mean is calculated across all questions in the dataset. The final FinScore combines the domain-normalized score with the hallucination penalty:
\begin{equation}\label{eq:finscore}
\mathcal{F} = F \cdot (1-P_H)\,,
\end{equation}
where $F$ is the domain-normalized average score across all questions and $P_H$ is the hallucination penalty rate. This multiplicative combination ensures that models are evaluated on both accuracy and reliability, with a strong emphasis on penalizing hallucination. In financial applications where incorrect predictions can lead to significant risks, models that hallucinate receive substantially lower scores regardless of their knowledge accuracy, reflecting the critical importance of reliable analysis.

\begin{table*}[t]
\centering
\renewcommand{\arraystretch}{1.1}
\setlength{\tabcolsep}{6pt}
\resizebox{\linewidth}{!}{%
\begin{tabular}{lccccccccc}
\toprule[1.2pt]
\rowcolor{gray!20}
\textbf{Method} & \textbf{Compre.} & \textbf{FG} & \textbf{Reason.} & \textbf{Single.} & \textbf{Multi.} & \textbf{Cal.} & \textbf{Avg.} & \textbf{FinScore} \\
\midrule[1pt]
\rowcolor{gray!10}
\multicolumn{9}{l}{\textit{Proprietary Models}} \\
\midrule[0.5pt]
Gemini Flash 2.0 & \textbf{49.89} & \textbf{59.07} & \textbf{48.71} & \textbf{63.73} & \textbf{54.11} & 35.59 & \textbf{51.85} & \textbf{20.10} \\
Claude 3.5 Sonnet & 45.99 & 55.28 & 43.35 & 59.61 & 47.59 & \textbf{37.35} & 48.20 & 15.61 \\
GPT-4o & 44.33 & 53.49 & 42.24 & 58.49 & 45.74 & 35.06 & 46.56 & 15.34 \\
DouBao-1.5V Pro & 44.42 & 54.33 & 43.48 & 58.55 & 47.36 & 35.43 & 47.26 & 15.03 \\
GPT-4o Mini & 41.91 & 48.47 & 42.88 & 52.38 & 45.42 & 31.27 & 43.72 & 11.70 \\
Claude 3.5 Haiku & 29.09 & 36.21 & 28.22 & 41.75 & 34.98 & 6.71 & 29.49 & 6.41 \\
\midrule[1pt]
\rowcolor{gray!10}
\multicolumn{9}{l}{\textit{Open-source Models}} \\
\midrule[0.5pt]
Qwen2.5-VL 72B & \textbf{49.64} & \textbf{60.25} & \textbf{49.44} & \textbf{65.06} & \textbf{54.26} & \textbf{36.60} & \textbf{52.54} & \textbf{20.87} \\
Qwen2-VL 72B & 37.11 & 51.68 & 33.92 & 58.05 & 36.81 & 32.77 & 41.72 & 11.50 \\
InternVL 2.5-8B & 37.96 & 51.83 & 35.33 & 59.43 & 38.60 & 28.24 & 41.90 & 10.42 \\
MiniCPM-O 2.6 & 37.71 & 53.17 & 35.98 & 60.21 & 39.05 & 30.31 & 42.74 & 9.77 \\
DeepSeekVL-2 & 32.91 & 51.46 & 29.63 & 60.41 & 35.73 & 18.33 & 38.08 & 8.28 \\
Qwen2-VL 7B & 34.14 & 48.17 & 31.73 & 54.88 & 35.07 & 26.32 & 41.80 & 6.91 \\
Qwen2.5-VL 3B & 32.53 & 52.55 & 30.70 & 61.29 & 31.98 & 30.15 & 39.87 & 6.95 \\
DeepSeekVL-2 Small & 34.14 & 51.00 & 31.55 & 59.73 & 34.81 & 26.85 & 38.18 & 6.11 \\
Phi-3 V & 27.52 & 45.59 & 26.97 & 54.35 & 26.57 & 25.73 & 34.45 & 3.87 \\
Phi-3.5 V & 25.73 & 43.37 & 26.46 & 51.84 & 24.24 & 27.12 & 33.13 & 2.85 \\
DeepSeekVL-2 Tiny & 23.06 & 31.48 & 21.14 & 37.14 & 25.97 & 7.88 & 24.45 & 2.05 \\
\bottomrule[1.2pt]
\end{tabular}%
}
\caption{\textbf{Performance Comparison} across different evaluation dimensions.}
\label{tab:main_results}
\end{table*}

\begin{table*}[t]
\centering
\renewcommand{\arraystretch}{1.2}
\setlength{\tabcolsep}{2pt}
\resizebox{\linewidth}{!}{%
\begin{tabular}{lcccccccccccccccc}
\toprule[1.2pt]
\rowcolor{gray!20}
\textbf{Method} & \textbf{Energy} & \textbf{Estate} & \textbf{Constr.} & \textbf{Metals} & \textbf{Chem.} & \textbf{Econo.} & \textbf{Asset} & \textbf{Fixed} & \textbf{Equity} & \textbf{Industrials} & \textbf{TMT} & \textbf{Trans.} & \textbf{General} & \textbf{Cons.} & \textbf{Pharma} & \textbf{Others} \\
\textbf{Method} & \textbf{Energy} & \textbf{Estate} & \textbf{Deriva.} & \textbf{Meteri.} & \textbf{Macroe.} & \textbf{Assets} & \textbf{Strate.} & \textbf{Fixed.} & \textbf{Equity} & \textbf{Indust.} & \textbf{TMT} & \textbf{Trans.} & \textbf{Financial} & \textbf{Consum.} & \textbf{Pharma} & \textbf{Others} \\
\midrule[1pt]
\rowcolor{gray!10}
\multicolumn{17}{l}{\textit{Proprietary Models}} \\
\midrule[0.5pt]
Gemini Flash 2.0 & \textbf{55.57} & \textbf{56.39} & \textbf{52.63} & \textbf{60.48} & \textbf{54.54} & 40.57 & \textbf{57.26} & 42.07 & \textbf{52.24} & \textbf{52.32} & \textbf{60.75} & \textbf{54.75} & \textbf{53.64} & \textbf{65.50} & \textbf{55.43} & \textbf{63.33} \\
Claude 3.5 Sonnet & 49.83 & 50.38 & 46.20 & 57.19 & 53.35 & 41.51 & 54.03 & 42.07 & 47.86 & 47.54 & 56.65 & 52.33 & 49.21 & 59.91 & 48.91 & 53.33 \\
GPT-4o & 50.00 & 52.63 & 47.95 & 57.19 & 52.57 & 41.51 & 54.03 & 42.07 & 46.03 & 44.02 & 51.93 & 48.88 & 49.04 & 57.81 & 45.59 & 60.00 \\
DouBao-1.5V Pro & 48.61 & 50.38 & 46.78 & 56.29 & 54.22 & \textbf{44.34} & 47.58 & \textbf{44.14} & 48.13 & 46.55 & 50.43 & 48.70 & 49.33 & 58.97 & 46.02 & 53.33 \\
GPT-4o Mini & 45.82 & 48.12 & 45.61 & 49.40 & 49.17 & 33.96 & 45.97 & 37.24 & 44.31 & 45.71 & 43.85 & 47.84 & 45.14 & 51.63 & 44.28 & 53.33 \\
Claude 3.5 Haiku & 29.79 & 32.33 & 33.92 & 35.33 & 40.49 & 32.08 & 40.32 & 32.41 & 31.96 & 30.94 & 34.29 & 39.21 & 29.30 & 33.92 & 28.22 & 36.67 \\
\midrule[1pt]
\rowcolor{gray!10}
\multicolumn{17}{l}{\textit{Open-source Models}} \\
\midrule[0.5pt]
Qwen-VL-2.5 72B & \textbf{56.79} & \textbf{57.14} & \textbf{51.46} & \textbf{60.78} & \textbf{58.17} & \textbf{44.34} & \textbf{54.84} & \textbf{38.62} & \textbf{53.05} & \textbf{53.87} & \textbf{58.01} & \textbf{55.27} & \textbf{54.57} & \textbf{64.80} & \textbf{57.16} & \textbf{66.67} \\
Qwen-VL-2 72B & 43.55 & 41.35 & 44.44 & 56.89 & 49.88 & 30.19 & 39.52 & 37.24 & 42.32 & 43.18 & 45.96 & 47.15 & 42.40 & 53.73 & 42.69 & 30.00 \\
InternVL-2.5 8B & 44.77 & 42.86 & 46.20 & 50.30 & 47.20 & 37.74 & 51.61 & 37.24 & 43.60 & 41.77 & 46.83 & 45.08 & 44.85 & 51.52 & 47.03 & 56.67 \\
MiniCPM-O 2.6 & 47.21 & 49.62 & 43.27 & 50.00 & 47.43 & 35.85 & 49.19 & \textbf{38.62} & 44.18 & 42.05 & 47.33 & 45.94 & 47.00 & 55.36 & 44.86 & 43.33 \\
DeepSeekVL-2 & 39.72 & 47.37 & 42.69 & 47.01 & 43.80 & 28.30 & 50.00 & 37.93 & 41.92 & 42.48 & 45.84 & 42.66 & 41.29 & 49.53 & 41.82 & 53.33 \\
Qwen-VL-2 7B & 45.99 & 42.11 & 41.52 & 44.01 & 42.38 & 30.19 & 41.13 & 34.48 & 38.91 & 42.05 & 45.09 & 40.93 & 41.35 & 48.72 & 41.24 & 40.00 \\
Qwen-VL-2.5 3B & 40.94 & 48.87 & 39.77 & 50.90 & 45.46 & 29.25 & 44.35 & 37.24 & 41.82 & 41.49 & 47.20 & 40.93 & 42.92 & 51.98 & 44.57 & 43.33 \\
DeepSeekVL-2 Small & 45.99 & 48.12 & 45.03 & 51.20 & 45.15 & 35.85 & 51.61 & 33.79 & 42.73 & 44.59 & 47.20 & 40.59 & 42.17 & 51.40 & 40.52 & 40.00 \\
Phi-3 V & 36.76 & 39.10 & 40.94 & 43.11 & 42.07 & 27.36 & 43.55 & 31.03 & 35.67 & 36.85 & 38.14 & 37.31 & 36.92 & 42.66 & 37.05 & 46.67 \\
Phi-3.5 V & 34.15 & 41.35 & 34.50 & 40.42 & 40.17 & 32.08 & 38.71 & 33.79 & 33.72 & 33.47 & 37.02 & 37.82 & 35.06 & 42.07 & 34.01 & 46.67 \\
DeepSeekVL-2 Tiny & 31.53 & 30.08 & 26.90 & 29.64 & 28.89 & 19.81 & 36.29 & 27.59 & 27.57 & 26.44 & 27.20 & 27.29 & 24.64 & 29.95 & 25.47 & 43.33 \\
\bottomrule[1.2pt]
\end{tabular}%
}
\caption{\textbf{Domain-specific Performance Comparison} across different sectors and industries.}
\label{tab:domain_results}
\end{table*}

\section{Experiment}\label{sec:experiment}

\subsection{Competing MLLMs}

To comprehensively evaluate the performance of current multimodal large language models in the financial domain, we conducted experiments across a diverse range of model architectures and parameter scales. Our evaluation encompasses both proprietary and open-source models. 
The proprietary models include GPT4o\footnote{https://openai.com/index/hello-gpt-4o/}, GPT4o-mini, Gemini Flash 2.0~\cite{team2023gemini}, Claude 3.5 Sonnet\footnote{https://www.anthropic.com/news/claude-3-5-sonnet}, 
 Claude 3.5 Haiku\footnote{https://www.anthropic.com/claude/haiku} and Doubao-1.5V Pro\footnote{https://team.doubao.com/zh/special/doubao\_1\_5\_pro}. For open-source alternatives, we selected Qwen2.5 VL 72B~\cite{yang2024qwen2}, 
InternVL 25-8B\footnote{https://internvl.opengvlab.com/},MiniCPM-O26~\cite{hu2024minicpmunveilingpotentialsmall}, 
DeepSeekVL-2~\cite{wu2024deepseekvl2mixtureofexpertsvisionlanguagemodels}, Qwen-2-VL-72B~\cite{Qwen2VL}, Qwen-2-VL-7B~\cite{Qwen2VL}, DeepseekVL-2 Small~\cite{wu2024deepseekvl2mixtureofexpertsvisionlanguagemodels}, Phi-3 128K~\cite{abdin2024phi3technicalreporthighly}, Phi-3.5 V~\cite{abdin2024phi3technicalreporthighly} and DeepSeekVL-2 Tiny~\cite{wu2024deepseekvl2mixtureofexpertsvisionlanguagemodels}.

\subsection{Evaluation Methods}
Our experimental evaluation was conducted separately for proprietary and open-source models. Proprietary models and larger open-source models were evaluated through commercial API calls, while smaller open-source models were deployed locally. All local experiments were performed on a single NVIDIA H100-level GPU. We utilized vLLM for efficient local deployment and inference.

\subsection{Main Results and Key Insights}

\paratitle{Proprietary Models' Performance.} Proprietary models demonstrate superior performance, with Gemini Flash 2.0 leading at average score and FinScore. The performance gap between proprietary and open-source models is most pronounced in multi-turn reasoning tasks.

\paratitle{Open-source Models' Performance.} Qwen2.5-VL 72B achieves competitive performance comparable to proprietary models, particularly excelling in fine-grained perception and single-turn tasks.

\paratitle{Task-Specific Performance.} All models perform better in single-turn tasks compared to multi-turn reasoning, with an average performance gap of 20-25\%. Calculation questions remain the most challenging dimension, with even top models achieving below 40\% accuracy.

\paratitle{Financial Domain Adaptation.} FinScore reveals significant gaps in financial domain expertise, with most open-source models scoring below 12, indicating room for improvement in financial knowledge and hallucination control.

\begin{figure}[t]
    \centering
    \includegraphics[width=\linewidth]{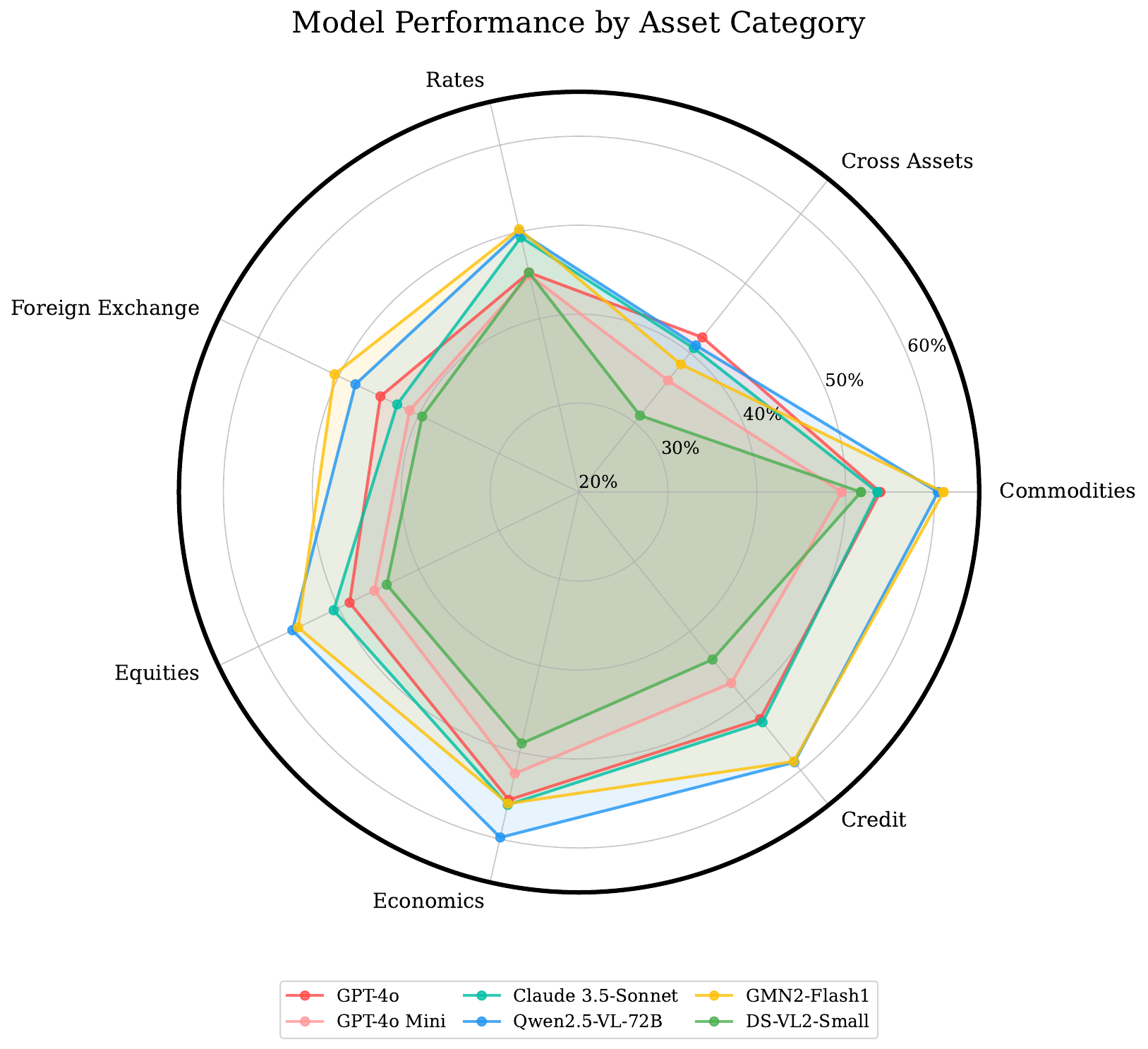}
    \caption{The radar chart of the asset class distribution of the dataset. }
    \label{fig:asset_radar}
\end{figure}

\subsection{Domain-specific Performance Analysis}

Through performance evaluation across 16 different industry domains, we observe significant variations in model capabilities. In traditional industrial sectors such as pharmaceuticals, energy, and metals, models generally demonstrate strong performance~(Gemini Flash 2.0), and energy and metals sectors consistently maintain scores above 50. However, economics and fixed income sectors present significant challenges, with even top models scoring below 45 points, indicating persistent difficulties in complex financial reasoning tasks. Notably, while smaller models consistently underperform across all domains, open-source models such as Qwen-VL-2.5 72B demonstrate competitive performance against proprietary models in specific domains, particularly in energy and metals. These findings not only reveal the current importance of model scale for domain expertise but also suggest promising developments in open-source models' ability to handle specialized tasks.

\subsection{Asset Class Analysis}
Analysis of the asset class distribution radar chart reveals notable performance variations across financial asset types. Models demonstrate strongest performance in the Commodities sector, followed by moderate performance in Credit and Rates categories. However, models show relatively weaker performance in the Foreign Exchange and Economics domains.

Notably, GPT4o and Claude 3.5-Sonnet exhibit robust overall capabilities across most asset classes. In contrast, smaller-scale models show acceptable performance only in specific categories like Commodities, while demonstrating lower overall effectiveness. These findings highlight the persistent disparities in multimodal large language models' comprehension capabilities within the financial domain, particularly in more complex areas like Foreign Exchange and Economics, indicating substantial room for improvement.

\subsection{Stability Analysis}

To assess the robustness and reliability of our evaluation framework, we conducted multiple rounds of testing and analyzed the standard deviations of model performance across different dimensions. As shown in Table~\ref{tab:std_results}, both GPT-4o and Qwen2-VL 7B demonstrate remarkable stability in their performance. The standard deviations across all evaluation dimensions remain consistently below 1\%, with GPT-4o showing variations between 0.58\% and 0.93\%, and Qwen2-VL 7B exhibiting fluctuations between 0.57\% and 0.86\%. These low variance levels indicate the high reliability and reproducibility of our evaluation framework, while also confirming the consistency of model behaviors across multiple test runs. The consistently low standard deviations across different model scales further validate the robustness of our evaluation methodology and the quality of our dataset.

\begin{table}[t]
\centering
\renewcommand{\arraystretch}{1.1}
\setlength{\tabcolsep}{4pt}
\resizebox{\linewidth}{!}{%
\begin{tabular}{lcccc}
\toprule[1.2pt]
\textbf{Method} & \textbf{Single.} & \textbf{Multi.} & \textbf{Cal.} & \textbf{Avg.} \\
\midrule[1pt]
GPT-4o & 58.49$\pm$0.93 & 45.74$\pm$0.77 & 35.06$\pm$0.58 & 46.56$\pm$0.64 \\
Qwen2-VL 7B & 58.05$\pm$0.85 & 36.81$\pm$0.86 & 32.77$\pm$0.62 & 41.72$\pm$0.57 \\
\bottomrule[1.2pt]
\end{tabular}%
}
\caption{\textbf{Model Performance with Standard Deviations} with $5$ runs.}
\vspace{-4mm}
\label{tab:std_results}
\end{table}



\section{Conclusion}\label{sec:conclusion}

This paper introduces \method{}, a comprehensive multimodal evaluation framework for the financial domain, comprising high-quality samples across 18 core financial domains. Our experiments demonstrate that leading MLLMs achieve unsatisfactory performance on \method{}, highlighting significant room for improvement in financial applications. The proposed FinScore metric, incorporating hallucination penalties and domain-normalized scoring, provides a robust evaluation framework for financial tasks, while maintaining prediction stability with low standard deviations across different prompts. Future work will focus on expanding dataset coverage, enhancing evaluation metrics, and promoting \method{}'s application in real-world financial analysis scenarios.

\section*{Limitations}

Despite \method{}'s carefully curated nature and substantial sample size, we acknowledge several limitations. Our evaluation methodology relies primarily on multiple-choice questions and calculations, which enables objective assessment but may not fully capture the complexity of real-world financial analysis tasks. Complex financial concepts posed interpretation difficulties even for knowledgeable annotators, potentially introducing subtle biases despite our quality control protocols. While \method{} covers diverse financial domains, it may not capture all scenarios encountered in financial work due to the vast and evolving nature of the industry, and currently lacks integration with audio/video content and real-time data analysis. Finally, although our stability analysis demonstrates robustness with high-quality inputs, these findings may not generalize to noisy or distorted inputs, highlighting that robustness to perturbations represents an important research direction building upon \method{}.

\section*{Acknowledgments}

This paper is partially supported by grants from the National Key Research and Development Program of China with Grant No. 2023YFC3341203, the National Natural Science Foundation of China (NSFC Grant Number 62276002), HKUST Start-up Fund (R9911), Theme-based Research Scheme grant (No.T45-205/21-N), InnoHK funding for Hong Kong Generative AI Research and Development Center, Hong Kong SAR. The authors are grateful to the anonymous reviewers for their efforts and insightful suggestions to improve this article.



\bibliography{custom}



\appendix
\label{appendix}

\section{More Related Work}\label{appendix:related_work}

While traditional multimodal benchmarks focused on specific tasks like captioning~\cite{chen2015microsoft, plummer2015flickr30k}, VQA~\cite{hudson2019gqa, goyal2017making, bigham2010vizwiz}, and specialized capabilities~\cite{sidorov2020textcaps, yang2021tap, li2019visualbert}, financial datasets such as ConvFinQA~\cite{chen2022convfinqa}, FINANCEBENCH~\cite{islam2023financebench}, FinBen~\cite{xie2024finbenholisticfinancialbenchmark}, CFBenchmark~\cite{lei2023cfbenchmark}, FinTextQA~\cite{chen2024fintextqadatasetlongformfinancial} and MME-Finance~\cite{gan2024mme}, FinVQA~\cite{bhatia2024fintral} either focus solely on language models or provide limited coverage of multimodal financial tasks, highlighting the need for comprehensive financial multimodal evaluation frameworks.

\section{Expert Consultation Process Record}\label{appendix:expert_consultation}
Our research design and validation process was strengthened through extensive consultation with financial industry experts. Through in-depth interviews, we gained valuable insights into real-world financial analysis workflows and information consumption patterns, which directly informed the design of \method{}. The expert panel included diverse professionals from investment banking, hedge funds, and asset management, with experience ranging from 5 to 10+ years:

\begin{itemize}
    \item [A] Investment Banking Professional with 5+ years of experience in ECM (Equity Capital Markets) and primary market equity issuance
    \item [B] Hedge Fund Sector Analyst with 10+ years of experience in industry research, specializing in new energy sectors
    \item [C] Hedge Fund Industry Researcher with 5+ years of experience
    \item [D] Investment Banking Professional with 5 years of experience in strategic equity and derivatives
    \item [E] Hedge Fund Industry Researcher with 10 years of experience
    \item [F] Asset Management Fund Manager with 5 years of experience
\end{itemize}

Key findings from our expert consultations highlighted several critical aspects that shaped our dataset design:

\textbf{Information Hierarchy and Consumption Patterns:} Experts consistently emphasized the importance of structured information access, typically beginning with executive summaries and investment views before diving into specific areas of interest. This insight directly influenced our hierarchical annotation structure in \method{}.

\textbf{Visual Data Interpretation:} Financial professionals heavily rely on charts and visualizations for trend analysis and comparative studies. Expert A and E particularly noted that visual representations often provide more intuitive insights than textual information, supporting our focus on diverse chart types and comprehensive visual analysis tasks.

\textbf{Multi-source Validation:} Expert C highlighted the practice of cross-referencing multiple sources and independently verifying data, emphasizing the importance of accuracy in financial analysis. This insight reinforced our rigorous quality control mechanisms and the inclusion of hallucination penalties in our evaluation metrics.

\textbf{Domain-specific Requirements:} Experts B and F emphasized the critical role of industry-specific knowledge and policy understanding, validating our approach to include comprehensive coverage across multiple financial domains and asset classes.

\textbf{Report Quality Variation:} Multiple experts noted significant variations in report quality across different sources, particularly between domestic and international research reports. This observation supported our decision to implement strict quality control measures and expert validation processes.

These expert insights were instrumental in developing \method{}'s comprehensive structure, ensuring its relevance to real-world financial analysis needs while maintaining high standards of quality and reliability. The consultation process validated our approach to creating a benchmark that effectively evaluates MLLMs' capabilities in handling complex financial tasks.

\section{Dataset Details}\label{appendix:dataset}

Our dataset organizes financial charts into 10 main categories, each with specific subcategories to facilitate precise classification and analysis. The main categories include Distribution Charts, Financial Charts, Flow Charts, Geographical Charts, Line Charts, Market Structure Charts, Proportional Charts, Relational Charts, Risk Distribution Charts, and Others. Each main category is further divided into specialized subcategories that capture specific visualization techniques and purposes. For example, Distribution Charts include histograms, box plots, and violin plots, while Financial Charts encompass line charts, K-line charts, and area charts. This hierarchical organization enables systematic evaluation of models' capabilities across different visualization types while maintaining clear categorization of financial data representation methods.

\section{Additional Results}\label{appendix:additional_results}

The experimental results demonstrate significant variations in model performance across different knowledge domains in financial analysis. Qwen25vl72b emerged as the leading performer, achieving exceptional scores particularly in consumer sectors and other specialized categories, suggesting that its architectural design and training approach are particularly well-suited for financial multimodal tasks. This performance advantage persisted across multiple domains, indicating robust and generalizable capabilities.

Notably, model size did not consistently correlate with performance effectiveness. This suggests that architectural choices and training strategies may be more crucial than raw model size for financial analysis tasks.

Domain complexity emerged as a significant factor in model performance patterns. Models generally excelled in sectors requiring straightforward analysis, such as consumer goods and TMT sectors, where performance consistently exceeded 50\% across leading models. However, significant challenges were observed in complex domains like broad asset allocation and strategy research, where most models struggled to achieve scores above 45\%. This performance gap highlights the increasing difficulty models face when dealing with multi-factor analysis and complex financial reasoning.

These findings carry important implications for the future development of multimodal models in finance. The success of specialized architectures like Qwen25vl72b suggests that domain-specific optimization may be more valuable than pursuing larger model sizes. Future research should focus on improving model performance in complex analytical domains while maintaining the strong performance observed in straightforward tasks. Additionally, the results emphasize the need for balanced capabilities across different financial sectors, particularly in areas requiring sophisticated reasoning and multi-factor analysis.

\section{Dataset Comparison}\label{appendix:dataset_comparison}

\begin{figure*}[t]
    \centering
    \includegraphics[width=\textwidth]{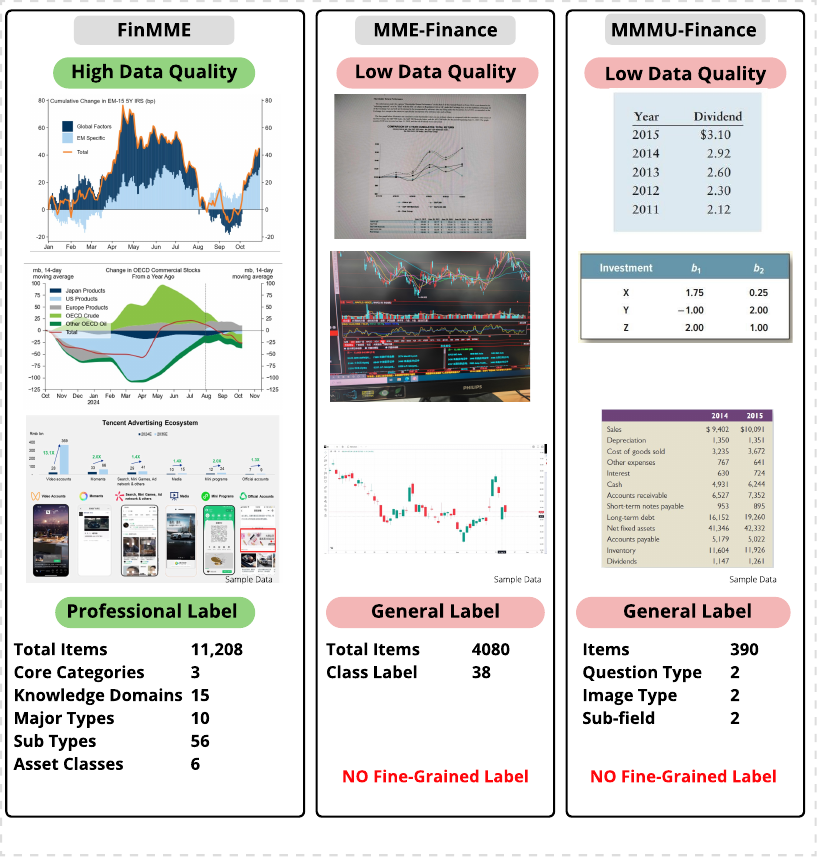}
    \caption{Data Comparison with related works.}
    \label{fig:sup:appendix_Dataset Compare}
\end{figure*}

As shown in Figure~\ref{fig:sup:appendix_Dataset Compare}, we compare \method{} with two other prominent financial multimodal datasets: MME-Finance and MMMU-Finance. The comparison reveals distinct characteristics and use cases for each dataset:

\textbf{\method{}} stands out with its high data quality and comprehensive coverage, containing more than 11,000 items across 3 core categories and 15 knowledge domains. It features professional-grade labeling with fine-grained annotations across 21 sub-types, providing detailed categorization of financial content. The dataset's comprehensive coverage spans 6 asset classes, establishing a structured hierarchy across multiple financial domains. A distinguishing aspect is its rigorous quality control system, implemented through expert validation processes that ensure the highest standards of financial accuracy and relevance.

\textbf{MME-Finance} offers a different focus with 4,080 items and 38 class labels. This dataset primarily emphasizes technical charts and trading data, making it particularly suited for market analysis applications. However, it employs general-purpose labeling without fine-grained annotations, resulting in less detailed categorization compared to \method{}. While it covers various financial aspects, its domain coverage is more limited, and the overall data quality is lower than \method{}, particularly in terms of annotation depth and expert validation.

\textbf{MMMU-Finance} is the most specialized of the three datasets, containing 390 items with a focused scope. It concentrates on fundamental business metrics such as sales, dividends, and investments, making it particularly relevant for corporate financial analysis. The dataset is structured around two primary question types and image types, with coverage limited to two sub-fields. Like MME-Finance, it employs general-purpose labeling without detailed annotations, which constrains its utility for complex financial analysis tasks.

This comparison highlights \method{}'s unique position in providing comprehensive, high-quality financial multimodal data with professional-grade annotations. While MME-Finance offers broader coverage of technical trading data and MMMU-Finance specializes in business metrics, \method{} delivers the depth and quality necessary for advanced financial analysis and model evaluation across multiple domains and asset classes. The combination of extensive coverage, detailed annotations, and rigorous quality control makes \method{} particularly well-suited for developing and evaluating sophisticated financial analysis models.



\begin{figure*}[t]
    \centering
    \includegraphics[width=\textwidth]{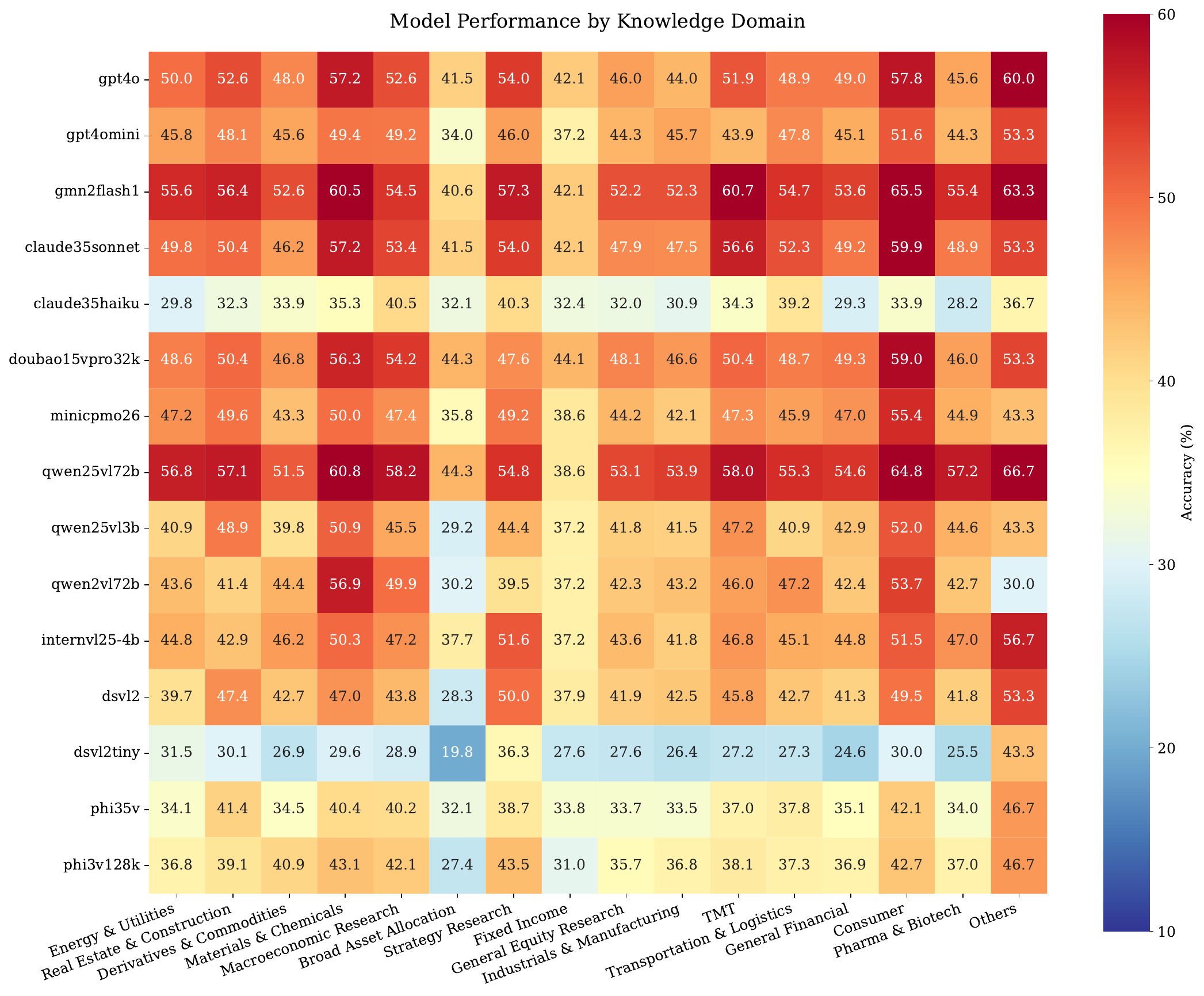}
    \caption{The heatmap of the knowledge domain distribution of the dataset. }
    \label{fig:sup:large_knowledge_domain_heatmap}
\end{figure*}

\end{document}